\documentclass[letterpaper]{article} 
\usepackage{aaai2026}  
\usepackage{times}  
\usepackage{helvet}  
\usepackage{courier}  
\usepackage[hyphens]{url}  
\usepackage{graphicx} 
\usepackage{natbib}  
\usepackage{caption} 
\usepackage{algorithm}
\usepackage{algorithmic}
\usepackage{multirow}
\usepackage{amssymb}
\usepackage{amsmath} 
\usepackage{booktabs}
\usepackage{newfloat}
\usepackage{listings}
\DeclareCaptionStyle{ruled}{labelfont=normalfont,labelsep=colon,strut=off} 
\floatstyle{ruled}
\newfloat{listing}{tb}{lst}{}
\floatname{listing}{Listing}
\nocopyright
\title{Stationarity Exploration for Multivariate Time Series Forecasting}
	
\author{
	Hao Liu\textsuperscript{\rm 1}
	Chun Yang\textsuperscript{\rm 1},
	Zhang xiaoxing\textsuperscript{\rm 2},
	Rui Ma \textsuperscript{\rm 1}
	Xiaobin Zhu \textsuperscript{\rm 1}
}

\affiliations{
	\textsuperscript{\rm 1}University of Science and Technology Beijing \\
	\textsuperscript{\rm 2}Yizhi, China Telecom\\
	d202410441@xs.ustb.edu.cn, chunyang@ustb.edu.cn,zhangxx7@chinatelecom.cn, m202421056@xs.ustb.edu.cn, xiaobinzhu@ustb.edu.cn
}

\usepackage{bibentry}

\begin{document}

\maketitle

\begin{abstract}
Deep learning-based time series forecasting has found widespread applications. Recently, converting time series data into the frequency domain for forecasting has become popular for accurately exploring periodic patterns. However, existing methods often cannot effectively explore stationary information from complex intertwined frequency components. In this paper, we propose a simple yet effective Amplitude-Phase Reconstruct Network (APRNet) that models the inter-relationships of amplitude and phase, which prevents the amplitude and phase from being constrained by different physical quantities, thereby decoupling the distinct characteristics of signals for capturing stationary information. Specifically, we represent the multivariate time series input across sequence and channel dimensions, highlighting the correlation between amplitude and phase at multiple interaction frequencies. We propose a novel Kolmogorov-Arnold-Network-based Local Correlation (KLC) module to adaptively fit local functions using univariate functions, enabling more flexible characterization of stationary features across different amplitudes and phases. This significantly enhances the model’s capability to capture time-varying patterns. Extensive experiments demonstrate the superiority of our APRNet against the state-of-the-arts (SOTAs).
\end{abstract}


\section{Introduction}
Time-series forecasting~\cite{arima,ACMsurvey,autoformer,Informer} is widely applied in fields such as finance, transportation, and the Internet of Things (IoT). Long-term forecasting~\cite{long1, long2, Are} requires considering a larger amount of historical data, more complex dynamic changes, and a greater number of potential influencing factors, making it highly valuable for practical applications.

Recent studies~\cite{fedformer,useful,timekan} transforming the time domain into the frequency domain~\cite{wavelet1,Fourier1,Fourier2} and selectively retaining the primary frequency components through filtering can effectively mitigate the impact of noise and outliers. Yi et al.~\cite{filternet} proposed frequency-aware shaping filters to enhance full-spectrum information utilization, while Ye et al.~\cite{non} developed a Fourier-based instance normalization method with pattern evolution prediction for non-stationary time series forecasting. Although the frequency domain effectively reveals key information that is difficult to directly observe in the time domain, real-world signals exhibit multi-frequency and interactive spectral characteristics. Existing methods~\cite{revin,autoformer,timekan,fedformer} struggle to simultaneously stationarize multi-frequency intertwined signals.

\begin{figure*}[h]
	\centering
	\includegraphics[width=\linewidth]{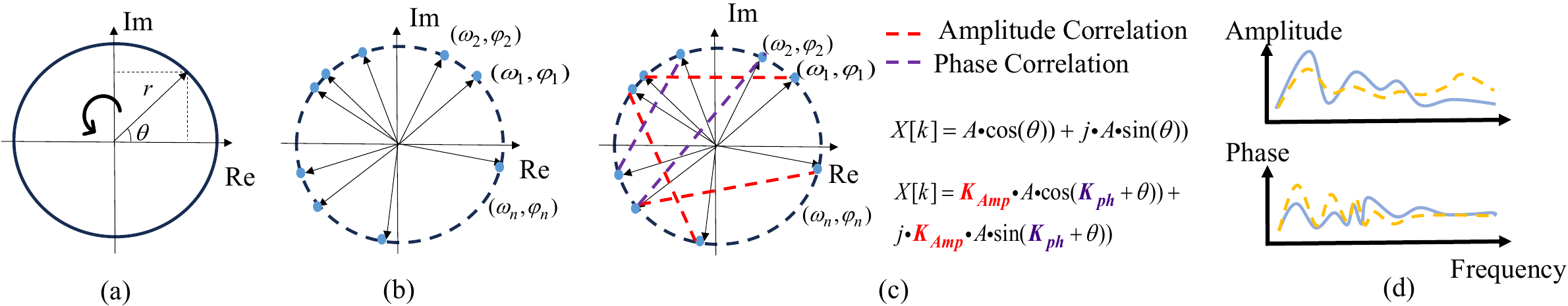}
	\caption{(a) The essence of the continuous Fourier transform is to decompose a time domain signal into a sum of components representing uniform circular motion with different frequencies on the complex unit circle. (b) Discrete sampling of the time series leads to a discrete Fourier transform, which yields a discrete spectrum. (c) The original temporal continuity is compromised, making it difficult to directly establish time-dependent relationships between sampled points. (d) The solid line represents the original frequency signal, while the dashed line represents the reconstructed frequency signal.}
	\label{principle}
\end{figure*}

To address the above-mentioned issue, we decompose the frequency components for observation from the complex plane. The decomposition of a signal via complex exponential basis functions through rotational operations is demonstrated in Fig.\ref{principle}(a), where each frequency component manifests as circular motion in the complex plane. In practical applications (Fig.\ref{principle}(b)), observed time series arise from the superposition of a finite number of discrete frequency components. Notably, the unit circle representation in the frequency domain reveals inherent non-stationary characteristics, as evidenced by the time-varying spectral properties. We propose the Amplitude-Phase Reconstruct Network (APRNet), a novel framework that decouples data along channel and sequence dimensions, enabling fine-grained modeling of cross-frequency relationships. As illustrated in Fig.~\ref{principle}(c), APRNet explicitly models amplitude and phase correlations between discrete signals. This process learning amplitude adjustment coefficients ($K_{Amp}$) and phase shifts ($K_{Ph}$) across different frequencies, the framework dynamically adjusts frequency domain signals to enhance stationarity-aware feature extraction, as shown in Fig.~\ref{principle}(d).

Stationary signals require time-invariant statistical properties. To achieve this, we leverage the smoothness and approximation capabilities of Kolmogorov-Arnold Network's (KAN)~\cite{kan} spline-based activation functions to preserve key frequency domain patterns during denoising. This approach fits the energy distribution of amplitudes and the time-frequency characteristics of phases. Amplitude adjustment ensures energy consistency, while phase adjustment maintains the temporal relationships of harmonics. Such a dual modulation mechanism effectively extracts stationary information across multiple frequency bands, significantly enhancing the model’s ability to analyze complex temporal patterns. Our main contributions are three-fold:

\begin{itemize}
	\item We propose a novel Amplitude-Phase Reconstruction Network (APRNet) to model the interrelationship between amplitude and phase from the frequency domain, efficiently decoupling signal characteristics to capture stationary information.
	
	\item We propose a novel KAN-based Local Correlation (KLC) module to fit of local stationary information using multiple sets of univariate functions for explicit representation, thereby substantially improving the model's capacity to capture time-varying patterns.
	
	\item Extensive experiments verify the superior performance and efficiency of our APRNet.
\end{itemize}

\section{Related Work}
\subsection{Deep learning-based Time Series Forecasting}
Early deep learning-based approaches for time series forecasting~\cite{dernn, rnn} primarily focused on developing adaptive and transferable deep model architectures. These efforts aimed to overcome the limitations of traditional statistical methods, which struggled to capture complex nonlinear patterns and long-term dependencies~\cite{longd1}. In this way, Informer~\cite{Informer} reduces computational complexity through the ProbSparse self-attention mechanism~\cite{transformer}, compresses sequence lengths via self-attention distillation, and employs a generative decoder for one-step long-sequence predictions, thus improving both efficiency and accuracy in long-term forecasting. Liu et al.~\cite{itransformer} treated each variable as an independent token to capture cross-variable dependencies in multivariate time series. However, transformer-based methods face challenges due to their large number of parameters and high memory consumption. As DL-based algorithms continue to evolve~\cite{pgn, diffusion1, koopa}, integrating diverse learning strategies has significantly enhanced model performance in time series analysis.
\subsection{Frequency Domain Forecasting}
In recent years, time series forecasting~\cite{useful, nonstationary, waveformer} has increasingly shifted toward frequency domain analysis to uncover key information that is difficult to capture in the time domain, making frequency domain analysis an emerging research focus. FLDmamba~\cite{fldmamba} leverages the Mamba state-space model to analyze Fourier and Laplace transforms, effectively capturing multi-scale periodicity and enhancing the model's robustness to data noise. FEDFormer~\cite{fedformer} introduces frequency domain processing and seasonal trend decomposition~\cite{autoformer} within the Transformer architecture~\cite{transformer}, reducing computational complexity to linear time through frequency domain random sampling. FourierGNN~\cite{fouriergnn} utilizes graph neural networks (GNN) to model spatiotemporal dependencies in multivariate time series and designs low-complexity graph operators in the Fourier space for efficient computation. Yi et al.~\cite{filternet} introduce filtering into time series forecasting with FilterNet, selectively transmitting or attenuating specific components of the time series signal to extract key informational patterns. However, the aforementioned studies showed weakness in stationary analysis in the frequency domain, which makes it difficult to handle complex patterns with multiple interacting frequency signals.
\begin{figure*}[h]
	\centering
	\includegraphics[width=\linewidth]{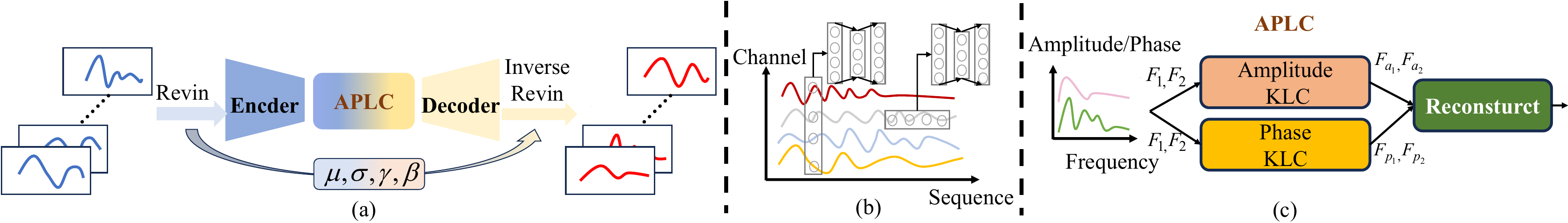}
	\caption{(a) The overview of our APRNet. (b) We perform separate modeling for channels and sequences, refine local dependencies, and jointly adjust signal relationships to obtain accurate prediction results.(c) We present the structure of the APLC module to reconstruct frequency signals.}
	\label{Overview}
\end{figure*}	
\subsection{Stationary Information Exploration}
The mining of locally stationary information involves transforming globally unpredictable non-stationary sequences into multiple locally predictable stationary segments, thereby unlocking the potential of the model. DLinear~\cite{Are} employs moving averages to separate seasonal and trend components, breaking down non-stationary time series into locally stationary components. S2IP-LLM~\cite{S2IP-LLM} utilizes large language models to analyze these decomposed components, enhancing the interpretability of forecasts. TimeBridge~\cite{timebridge} segments the input sequence into multiple patches and applies patch-wise attention to each patch, capturing stable dependency relationships among each variable. TimeMixer~\cite{timexer} follows a fine-to-coarse principle to decompose features across multiple scales and further analyzes the stationary information. Building on this. TimeMixer++~\cite{TimeXier} defines various resolutions in the frequency domain and adopts diverse hybrid strategies to extract complex, task-adaptive time series patterns. Inspired by these studies, we explicitly model multiple intertwined frequencies in frequency domain via amplitude adjustments and phase shifts, thereby accurately simulate complex patterns in time series forecasting.


\section{Methodology}
\subsection{Amplitude-Phase Reconstruct Network}
The architecture of the Amplitude-Phase Network (APRNet) is shown in Fig.~\ref{Overview}(a), which consists primarily of the Revin, Encoder, Decoder, and the \textbf{APLC} module. Specifically, for a given time series \( X = \{x^{(1)}_{1:T}, x^{(2)}_{1:T}, \dots, x^{(n)}_{1:T}\} \), where \( T \) is the sequence length and the superscript denotes the sequence number, we define the channel dimension of sequence \( x \) as \( C \), with \( x^{(n)}_{1:T} \in \mathbb{R}^{N \times C} \) and \( X \in \mathbb{R}^{B \times N \times C} \).

To address the issue of non-stationary sequences, where the distribution of the predictive model changes over time, we employ Revin~\cite{revin}. This prevents performance degradation during testing due to covariate or conditional shifts. Initially, we apply mean and standard deviation normalization to the data sequence \( X \)~\cite{reverse,non-stationary}, which can be formulated as:
\begin{equation}
	\overline{X} = \mathbf{Norm}(X) = \frac{X - \text{Mean}(x)}{\text{Std}(X) + \delta},
\end{equation}
where \( \overline{X} \in \mathbb{R}^{B \times N \times C} \). Revin alleviates the difficulties posed by non-stationarity through the following transformation:
\begin{equation}
	\hat{X} = \mathbf{Revin}(\overline{X}) = \gamma \overline{X} + \beta, \label{eq:5}
\end{equation}
where \( \hat{X} \in \mathbb{R}^{B \times N \times C} \), \( \gamma \) and \( \beta \) are learnable parameters. 
Then, we use an Encoder model to map temporal data to the latent space. This can be formulated as:
\begin{equation}
	Z = \mathbf{LayerNorm}(\mathbf{f}(\hat{X})),
\end{equation}
where \( \mathbf{f} \) is a linear layer that maps the feature dimension from \( C \) to \( K \).

For frequency domain information, we perform local modeling within an \textbf{APLC} model, capturing both Amplitude and Phase dependencies of the frequency domain features. The details of the \textbf{APLC} module are in the next section. We believe that the output of the \textbf{APLC} is a frequency domain representation, and it requires a decoding module to learn temporal dependencies, which are modeled separately via individual linear layers:
\begin{equation}
	\overline{Y} = \mathbf{W_t} \cdot \mathbf{LayerNorm}(\hat{Y}),
\end{equation}
where \( \hat{Y} \) is the output of the \textbf{APLC} module.

Finally, the output \( \overline{Y} \) is transformed through Revin’s invertible transformation, preserving all non-stationary information from the original sequence to obtain the prediction result, which can be formulated as:
\begin{equation}
	Y = \mathbf{InverseRevin}(\mathbf{InverseNorm}(\overline{Y})),
\end{equation}
where $\overline{Y}\in \mathbb{R}^{B\times N\times \tau}$, $\tau$ is the forecast length, the output \( Y \in \{y^{(1)}_{t+T+1:t+T+\tau}, \dots, y^{(n)}_{t+T+1:t+T+\tau}\} \).

\subsection{Amplitude-Phase Local Correlation Module}
The structure of the \textbf{APLC} module is illustrated in Fig.~\ref{Overview} (b). To efficiently extract stationary information, we perform joint modeling of the input from both the sequence and channel dimensions to capture fine-grained frequency domain signals dependencies, and prevent the loss of channel information to obtain entangled and reliable signal relationships. As shown in Fig.~\ref{Overview} (b), we perform Fourier transforms on the temporal and channel dimensions separately to obtain global represents \( F_{1} \in \mathbb{R}^{B \times \left(\frac{N}{2}+1\right) \times K} \) and \( F_{2} \in \mathbb{R}^{B \times N \times \left(\frac{K}{2}+1\right)} \). Here, \( F_{1} \) denotes the sequence dimension of the frequency feature, and \( F_{2} \) denotes that of the frequency feature. In the frequency domain, we reshape the temporal dimension feature from \( F_{1} \in \mathbb{R}^{B \times \left(\frac{N}{2}+1\right) \times K} \) to \( F_{1} \in \mathbb{R}^{B \times K \times \left(\frac{N}{2}+1\right)} \).
Subsequently, we decouple the frequency domain features, capture the relationship between amplitude and phase through the APLC module, explore the stationary information, and reconstruct the frequency domain signal based on the stationary features.  

In addition, we designed a KAN-based Local Correlation (KLC) module to extract the local relationships between amplitude and phase.
KLC utilizes univariate functions to achieve adaptive fitting of local functions, obtaining the stationary characteristics of amplitude and phase for each frequency component, and subsequently derive corresponding reconstruction coefficients  $(F_{a_1}, F_{p_1} \in \mathbb{R}^{B \times K \times \left(\frac{N}{2}+1\right)})$ for amplitude and $(F_{a_2}, F_{p_2} \in \mathbb{R}^{B \times N \times \left(\frac{K}{2}+1\right)})$ for phase. The process of reconstruction and transformation to the time domain can be formulated as:
\begin{equation}
	\begin{split}
		K_{1} = &\mathbf{IFFT}\left(\hat{F}_{a_1} \odot F_{a_1} \cos(\hat{F}_{p_1} + F_{p_1}) \right. \\
		&\left. + j \left( \hat{F}_{a_1} \odot F_{a_1} \sin(\hat{F}_{p_1} + F_{p_1}) \right) \right), \label{eq:13}
	\end{split}
\end{equation}
where \( K_{1} \in \mathbb{R}^{B \times N \times K }\) represents the reconstructed time dimension feature, \( K_{2} \in \mathbb{R}^{B \times K \times N} \) represents the reconstructed channel dimension feature and employs methods similar to used for the sequence. The \( \odot \) denotes element-wise multiplication.

Multiple frequency components overlap in the frequency domain, resulting in complex mixed statistical properties. We adjust the spectral amplitudes via element-wise multiplication to enhance the energy contrast between signals. Meanwhile, phase addition operations are employed to correct time-delay discrepancies, optimizing time-frequency alignment. This reconstruction process separates frequency components, thereby extracting local stationary features in the frequency domain of the channel and sequence.

Finally, We integrate relational features across both temporal and channel dimensions using learnable fusion parameters, which can be formulated as:
\begin{equation}
	\hat{Y} = Z + \alpha K_{1} + \beta K_{2},
\end{equation}
where \( \alpha \) and \( \beta \) are learnable parameters.

\subsection{KAN-based Local Correlation Module}
Inspired by both channel attention~\cite{senet} and self-attention mechanisms~\cite{transformer,S2IP-LLM}, we consider the intensity at corresponding frequencies in the amplitude spectrum and phase spectrum when constructing local frequency correspondences via KAN linear layers. Given that the sequence dimension feature \( F_{1} \) and the channel dimension feature \( F_{2} \) undergo the same process, we focus on introducing the sequence dimension feature. The frequency dependent scaling factors are computed as follows: 
\begin{equation}
	F_{local} = \phi_{ij}(F_{1_i})+b_j,
\end{equation}
where $\phi$ is a learnable function on the input-to-output path, and $b$ is the bias term. Each $\phi(\cdot)$ is implemented via B-spline curves, witch can be formulated as:
\begin{equation}
	\phi(x) = \sum_{k} c_k \cdot B_k(x),
\end{equation}
where $B(\cdot)$ is the B-spline basis function, and $c_k(\cdot)$ represents the learnable coefficients.
A linear layer followed by the Softmax function generates frequency dependent scaling factors based on these amplitude relationships, compensating the amplitude. This can be formulated as:
\begin{equation}
	\hat{F}_{a_1} = \sigma(\mathbf{f}_{align_{a_1}}(F_{local})),
\end{equation}
where \( \sigma \) is the Sigmoid activation function. Following the amplitude and phase KLC, we establish local correspondences between phases of different frequencies. In this case, we employ addition to optimize time-frequency alignment. 

\section{Experiment}
\noindent \textbf{Datasets.}
We utilize the ETT~\cite{Informer}, Weather~\cite{autoformer}, Electricity~\cite{autoformer}, and Traffic~\cite{autoformer} public datasets as benchmarks for long-term time series forecasting. The M4~\cite{m4} competition datasets are a benchmark for short-term forecasting, while the ETT datasets are used for zero-shot forecasting. Additionally, the SMD~\cite{SMD}, SML, SMAP~\cite{SMAP}, SWaT~\cite{swat}, and PSM~\cite{PSM} datasets are used for anomaly detection.

\noindent \textbf{Metrics.}
We follow current research standards in evaluating time series forecasting models. For long-term and zero-shot forecasting~\cite{aligning}, we use Mean Squared Error (MSE) and Mean Absolute Error (MAE) as evaluation metrics~\cite{TimeXier}. For short-term forecasting, we follow the metrics of SCINet~\cite{scinet} and adopt Symmetric Mean Absolute Percentage Error (SMAPE), Mean Absolute Scaled Error (MASE), and Overall Weighted Average (OWA).

\noindent \textbf{Baselines.}
In our experiments, we benchmark against current state-of-the-art (SOTA) methods for comparative analysis. In time series forecasting, the outstanding performer TimeMixer++ serves as our primary comparison algorithm~\cite{TimeXier}. The following algorithms are also included in our experiments: TQNet~\cite{timemoe}, LLM-TS~\cite{llmts}, CSFormer~\cite{csformer}, FilterNet~\cite{filternet}, CycleNet~\cite{cyclenet}, Time-LLM~\cite{Time-LLM}, iTransformer~\cite{itransformer}, DLinear~\cite{Are}, S2IP-LLM~\cite{S2IP-LLM}, and GPT4TS~\cite{GPT4TS}.

\subsection{Experiment Results}
\noindent \textbf{Long-Term Forecasting.} The detailed results are listed in Table~\ref{tab:long}. We conducted experiments with an input sequence length set to 512. The experimental results demonstrate that our APRNet achieves outstanding performance across seven publicly available datasets. Some of the state-of-the-art (SOTA) methods in the experiments lack experimental results because the original paper did not provide specific experimental outcomes for these algorithms, and upon our reproduction attempts, we found that key experimental parameters were missing. Overall, our method shows significant advantages over the current state-of-the-art (SOTA) methods. Specifically, on the ETTh1 dataset, although our average forecast horizon error is slightly higher than that of Time-MOE, our method outperforms in terms of comprehensive performance evaluation. On the ETTm2 dataset, while our method produces sub-optimal results at different time steps, its average forecasting result is the best. For the weather and electricity datasets, which feature a large number of samples and rich multivariate characteristics, our method achieves excellent performance. Compared to the sub-optimal results, our method reduces the average error by 2.2\% on the weather dataset and by 4.2\% on the electricity dataset. Moreover, our method also achieves optimal forecasting performance for the traffic dataset.
\begin{table*}[h]
	\centering
	\scalebox{0.630}{
		    \begin{tabular}{c|c|cc|cc|cc|cc|cc|cc|cc|cc|cc}
		    	\toprule
		    	& \multirow{2}[2]{*}{Models} & \multicolumn{2}{c|}{\multirow{2}[2]{*}{APRNet}} & \multicolumn{2}{c|}{TimeMixer++} & \multicolumn{2}{c|}{TQNet} & \multicolumn{2}{c|}{CSFormer} & \multicolumn{2}{c|}{LLM-TS} & \multicolumn{2}{c|}{CycleNet} & \multicolumn{2}{c|}{FilterNet} & \multicolumn{2}{c|}{Time-LLM} & \multicolumn{2}{c}{iTransformer} \\
		    	&       & \multicolumn{2}{c|}{} & \multicolumn{2}{c|}{2025 } & \multicolumn{2}{c|}{2025} & \multicolumn{2}{c|}{2025 } & \multicolumn{2}{c|}{2025 } & \multicolumn{2}{c|}{2024 } & \multicolumn{2}{c|}{2024 } & \multicolumn{2}{c|}{2024 } & \multicolumn{2}{c}{2024 } \\
		    	\midrule
		    	Datasets & Mertics & MSE   & MAE   & MSE   & MAE   & MSE   & MAE   & MSE   & MAE   & MSE   & MAE   & MSE   & MAE   & MSE   & MAE   & MSE   & MAE   & MSE   & MAE \\
		    	\multirow{5}[0]{*}{ETTh1} & 96    & \textbf{0.364 } & \underline{0.395 } & 0.386  & 0.414  & 0.379  & 0.403  & \underline{0.372 } & \textbf{0.394 } & 0.403  & 0.420  & 0.380  & 0.407  & 0.381  & 0.408  & 0.383  & 0.404  & 0.395  & 0.420  \\
		    	& 192   & \textbf{0.398 } & \textbf{0.417 } & 0.419  & 0.435  & 0.417  & 0.430  & 0.420  & \underline{0.425 } & 0.440  & 0.441  & 0.409  & 0.426  & \underline{0.412 } & 0.427  & 0.427  & 0.431  & 0.427  & 0.441  \\
		    	& 336   & \textbf{0.424 } & \textbf{0.435 } & 0.449  & 0.458  & 0.452  & 0.457  & 0.453  & 0.440  & 0.471  & 0.457  & \underline{0.430 } & 0.439  & 0.431  & 0.439  & \underline{0.430 } & \underline{0.436 } & 0.445  & 0.457  \\
		    	& 720   & 0.468  & 0.478  & 0.494  & 0.497  & 0.505  & 0.510  & 0.470  & 0.470  & 0.503  & 0.487  & 0.474  & 0.489  & \textbf{0.438 } & \textbf{0.454 } & \underline{0.465 } & \underline{0.469 } & 0.537  & 0.530  \\
		    	& Avg   & \textbf{0.414 } & \textbf{0.431 } & 0.437  & 0.451  & 0.438  & 0.450  & 0.429  & 0.432  & 0.454  & 0.451  & 0.423  & 0.440  & \underline{0.416 } & \underline{0.432 } & 0.426  & 0.435  & 0.451  & 0.462  \\
		    	\midrule
		    	\multirow{5}[1]{*}{ETTh2} & 96    & \textbf{0.279 } & \underline{0.342 } & 0.288  & 0.351  & \underline{0.287 } & 0.350  & 0.293  & \textbf{0.340 } & 0.322  & 0.366  & 0.294  & 0.352  & 0.300  & 0.360  & 0.293  & 0.348  & 0.304  & 0.360  \\
		    	& 192   & \textbf{0.341 } & \textbf{0.382 } & 0.362  & 0.395  & \underline{0.348 } & \underline{0.389 } & 0.375  & 0.390  & 0.400  & 0.409  & 0.354  & 0.391  & 0.369  & 0.404  & 0.356  & 0.391  & 0.377  & 0.403  \\
		    	& 336   & \textbf{0.372 } & \underline{0.411 } & 0.408  & 0.429  & 0.382  & \underline{0.412 } & \underline{0.378 } & \textbf{0.406 } & 0.432  & 0.435  & 0.381  & 0.416  & 0.389  & 0.428  & \textbf{0.372 } & \underline{0.408 } & 0.405  & 0.429  \\
		    	& 720   & \textbf{0.403 } & \underline{0.439 } & 0.432  & 0.453  & 0.415  & 0.445  & \underline{0.409 } & \textbf{0.432 } & 0.430  & 0.442  & 0.412  & 0.444  & 0.454  & 0.471  & 0.421  & 0.446  & 0.443  & 0.464  \\
		    	& Avg   & \textbf{0.349 } & \underline{0.394 } & 0.373  & 0.407  & \underline{0.358 } & 0.399  & 0.364  & \textbf{0.392 } & 0.396  & 0.413  & 0.360  & 0.401  & 0.378  & 0.416  & 0.361  & 0.398  & 0.382  & 0.414  \\
		    	\midrule
		    	\multirow{5}[2]{*}{ETTm1} & 96    & \underline{0.293 } & 0.347  & 0.298  & 0.355  & \textbf{0.291 } & \textbf{0.344 } & 0.324  & 0.367  & 0.329  & 0.371  & 0.297  & 0.353  & 0.318  & 0.358  & 0.294  & \underline{0.345 } & 0.312  & 0.366  \\
		    	& 192   & \underline{0.331 } & \underline{0.371 } & 0.336  & 0.379  & 0.335  & 0.373  & 0.369  & 0.388  & 0.380  & 0.398  & 0.340  & 0.374  & 0.338  & 0.378  & \textbf{0.330 } & \textbf{0.368 } & 0.347  & 0.385  \\
		    	& 336   & \textbf{0.363 } & \textbf{0.390 } & 0.381  & 0.403  & \underline{0.364 } & 0.395  & 0.396  & 0.408  & 0.418  & 0.426  & 0.372  & \underline{0.392 } & 0.368  & 0.395  & 0.365  & \underline{0.392 } & 0.379  & 0.404  \\
		    	& 720   & \underline{0.425 } & \underline{0.422 } & 0.434  & 0.429  & 0.429  & 0.433  & 0.451  & 0.439  & 0.476  & 0.440  & 0.433  & 0.424  & \textbf{0.403 } & \textbf{0.413 } & 0.427  & 0.431  & 0.441  & 0.442  \\
		    	& Avg   & \textbf{0.353 } & \textbf{0.383 } & 0.362  & 0.392  & 0.355  & 0.386  & 0.385  & 0.400  & 0.401  & 0.409  & 0.361  & 0.386  & 0.357  & 0.386  & \underline{0.354 } & \underline{0.384 } & 0.370  & 0.399  \\
		    	\midrule
		    	\multirow{5}[2]{*}{ETTm2} & 96    & \textbf{0.169 } & 0.261  & \textbf{0.169 } & \underline{0.258 } & 0.171  & 0.261  & \underline{0.170 } & 0.269  & 0.189  & 0.266  & 0.179  & 0.267  & 0.174  & \textbf{0.257 } & 0.175  & 0.265  & 0.179  & 0.271  \\
		    	& 192   & \textbf{0.224 } & \textbf{0.297 } & \textbf{0.224 } & \textbf{0.297 } & \underline{0.230 } & \underline{0.300 } & 0.244  & 0.309  & 0.253  & 0.307  & 0.235  & 0.307  & 0.240  & 0.330  & 0.243  & 0.316  & 0.242  & 0.313  \\
		    	& 336   & \textbf{0.273 } & \textbf{0.330 } & \underline{0.274 } & \underline{0.333 } & 0.283  & 0.338  & 0.303  & 0.346  & 0.315  & 0.345  & 0.278  & 0.336  & 0.297  & 0.339  & 0.294  & 0.343  & 0.288  & 0.344  \\
		    	& 720   & \textbf{0.353 } & \underline{0.382 } & 0.365  & 0.388 & \underline{0.360 } & \textbf{0.380 } & 0.400  & 0.400  & 0.421  & 0.408  & 0.385  & 0.389  & 0.362  & 0.389  & 0.389  & 0.410  & 0.378  & 0.397  \\
		    	& Avg   & \textbf{0.255 } & \textbf{0.318 } & \underline{0.258 } & \underline{0.319 } & 0.261  & 0.320  & 0.282  & 0.331  & 0.295  & 0.331  & 0.269  & 0.325  & 0.268  & 0.329  & 0.275  & 0.334  & 0.272  & 0.331  \\
		    	\midrule
		    	\multirow{5}[2]{*}{Weather} & 96    & \textbf{0.144 } & \textbf{0.196 } & \underline{0.148 } & 0.203  & 0.149  & \underline{0.202 } & 0.168  & 0.217  & 0.403  & 0.420  & 0.153  & 0.210  & 0.151  & 0.204  & 0.163  & 0.210  & 0.253  & 0.304  \\
		    	& 192   & \textbf{0.191 } & \textbf{0.243 } & \underline{0.194 } & \textbf{0.243 } & 0.201  & 0.249  & 0.213  & 0.257  & 0.440  & 0.441  & 0.196  & 0.248  & 0.198  & 0.248  & 0.205  & \underline{0.245 } & 0.280  & 0.319  \\
		    	& 336   & \textbf{0.238 } & \textbf{0.281 } & \underline{0.245 } & \underline{0.283 } & 0.247  & 0.287  & 0.272  & 0.298  & 0.471  & 0.457  & 0.251  & 0.290  & 0.247  & 0.285  & 0.257  & 0.287  & 0.321  & 0.344  \\
		    	& 720   & \textbf{0.315 } & \textbf{0.332 } & 0.319  & \underline{0.337 } & \underline{0.312 } & \textbf{0.332 } & 0.346  & 0.347  & 0.503  & 0.487  & 0.323  & 0.338  & 0.318  & 0.345  & 0.323  & \textbf{0.332 } & 0.364  & 0.374  \\
		    	& Avg   & \textbf{0.222 } & \textbf{0.263 } & \underline{0.227 } & \underline{0.267 } & \underline{0.227 } & 0.268  & 0.250  & 0.280  & 0.454  & 0.451  & 0.231  & 0.272  & 0.229  & 0.271  & 0.237  & 0.269  & 0.305  & 0.335  \\
		    	\midrule
		    	\multirow{5}[2]{*}{ECL} & 96    & \textbf{0.125 } & \textbf{0.220 } & 0.130  & 0.225  & 0.131  & 0.228  & 0.146  & 0.242  & 0.167  & 0.271  & 0.129  & \underline{0.224 } & \underline{0.127 } & \underline{0.224 } & 0.140  & 0.236  & 0.147  & 0.248  \\
		    	& 192   & \textbf{0.145 } & \textbf{0.240 } & 0.152  & 0.244  & 0.153  & 0.251  & 0.172  & 0.266  & 0.178  & 0.280  & \underline{0.146 } & \underline{0.242 } & 0.150  & 0.244  & 0.150  & 0.249  & 0.165  & 0.267  \\
		    	& 336   & \textbf{0.163 } & \textbf{0.260 } & 0.169  & \underline{0.262 } & 0.169  & 0.268  & 0.176  & 0.271  & 0.198  & 0.302  & \underline{0.165 } & \underline{0.262 } & 0.171  & 0.267  & 0.168  & 0.267  & 0.178  & 0.279  \\
		    	& 720   & \textbf{0.190 } & \textbf{0.285 } & 0.205  & 0.294  & \underline{0.192 } & \underline{0.287 } & 0.211  & 0.303  & 0.233  & 0.344  & 0.202  & 0.302  & 0.200  & 0.293  & 0.209  & 0.302  & 0.322  & 0.398  \\
		    	& Avg   & \textbf{0.156 } & \textbf{0.251 } & 0.164  & \underline{0.256 } & \underline{0.161 } & 0.259  & 0.176  & 0.270  & 0.173  & 0.266  & \underline{0.161 } & 0.258  & 0.162  & 0.257  & 0.167  & 0.264  & 0.203  & 0.298  \\
		    	\midrule
		    	\multirow{5}[2]{*}{Traffic} & 96    & \textbf{0.349 } & \textbf{0.253 } & 0.368  & \underline{0.266 } & \underline{0.366 } & 0.255  & ——    & ——    & 0.587  & 0.315  & 0.392  & 0.280  & 0.366  & \underline{0.266 } & 0.384  & 0.278  & 0.367  & 0.288  \\
		    	& 192   & \underline{0.379 } & \textbf{0.263 } & 0.381  & \underline{0.264 } & 0.392  & 0.276  & ——    & ——    & 0.612  & 0.326  & 0.402  & 0.279  & 0.384  & 0.275  & 0.398  & 0.286  & \textbf{0.378 } & 0.293  \\
		    	& 336   & \underline{0.394 } & \textbf{0.271 } & \textbf{0.393 } & \textbf{0.271 } & 0.400  & 0.280  & ——    & ——    & 0.634  & 0.338  & 0.414  & 0.289  & 0.395  & \underline{0.280}  & 0.408  & 0.289  & 0.399  & 0.294  \\
		    	& 720   & \underline{0.435 } & 0.298  & \underline{0.435 } & \textbf{0.292 } & 0.442  & 0.300  & ——    & ——    & 0.640  & 0.351  & 0.443  & \underline{0.295 } & \textbf{0.433 } & 0.299  & 0.436  & 0.303  & 0.442  & 0.304  \\
		    	& Avg   & \textbf{0.389 } & \textbf{0.271 } & \underline{0.394 } & \underline{0.273 } & 0.400  & 0.278  & ——    & ——    & 0.618  & 0.333  & 0.413  & 0.286  & 0.395  & 0.280  & 0.407  & 0.289  & 0.397  & 0.295  \\
		    	\bottomrule
		    \end{tabular}%
		
	}
	\caption{Long-term time series forecasting results, with an input sequence length of 512 and prediction lengths of $\{96, 192, 336, 720\}$. Avg text indicates mean value. \textbf{Bolded} represents the best results. \underline{Underline} indicates the second-best.}\label{tab:long}%
\end{table*}%

\noindent \textbf{Few-Shot Forecasting.} Few-shot forecasting requires extracting latent time-dependent patterns from minimal historical data, posing significant challenges in handling non-stationary dynamic changes and accurately measuring uncertainty. The exploration of local stationary information by APRNet gains insights into the periodic information of the entire dataset using a small amount of data. Compared with the current state-of-the-art (SOTA) methods, we achieved the best performance on the ETTh1 and ETTh2 datasets, as shown in Table~\ref{tab:few}. Notably, in the few-shot forecasting results on ETTh1, we reduced the error by 35.1\% compared to the second-best method. From an overall perspective, APRNet demonstrates superior performance.
\begin{table}[H]
	\centering
	\scalebox{0.650}{
		\begin{tabular}{c|c|cc|cc|cc|cc}
			\toprule
			&       & \multicolumn{2}{c|}{APRNet} & \multicolumn{2}{c|}{TimeMixer++} & \multicolumn{2}{c|}{Time-LLM} & \multicolumn{2}{c}{iTranformer}   \\
			\midrule
			&       & MSE   & MAE   & MSE   & MAE   & MSE   & MAE   & MSE   & MAE    \\
			\midrule
			\multirow{5}[2]{*}{ETTh1} & 96    & \textbf{0.377 } & \textbf{0.407 } & 0.715  & 0.588  & \underline{0.518 } & \underline{0.498 } & 0.808  & 0.610   \\
			& 192   & \textbf{0.415 } & \textbf{0.428 } & 0.728  & 0.589  & \underline{0.702 } & \underline{0.547 } & 0.928  & 0.658    \\
			& 336   & \textbf{0.467 } & \textbf{0.462 } & 0.742  & 0.592  & \underline{0.725 } & \underline{0.603 } & 1.475  & 0.861    \\
			& 720   & ——    & ——    & ——    & ——    & ——    & ——    & ——    & ——   \\
			& Avg   & \textbf{0.420 } & \textbf{0.432 } & 0.728  & 0.590  & \underline{0.648 } & \underline{0.549 } & 1.070  & 0.710   \\
			\midrule
			\multirow{5}[2]{*}{ETTh2} & 96    & \textbf{0.303 } & \textbf{0.343 } & 0.445  & 0.459  & 0.384  & 0.420  & 0.397  & 0.427  \\
			& 192   & \underline{0.396 } & \textbf{0.420 } & 0.523  & 0.512  & \textbf{0.394 } & \underline{0.424 } & 0.438  & 0.445   \\
			& 336   & \underline{0.428 } & \underline{0.440 } & 0.549  & 0.506  & \textbf{0.416 } & \textbf{0.433 } & 0.631  & 0.553  \\
			& 720   & ——    & ——    & ——    & ——    & ——    & ——    & ——    & ——    \\
			& Avg   & \textbf{0.376 } & \textbf{0.401 } & 0.506  & 0.492  & 0.398  & 0.426  & 0.488  & 0.475   \\
			\midrule
			\multirow{5}[2]{*}{ETTm1} & 96    & \textbf{0.414 } & \underline{0.425 } & 0.425  & 0.432  & \underline{0.422 } & \textbf{0.424 } & 0.589  & 0.510    \\
			& 192   & \textbf{0.446 } & \underline{0.446 } & 0.513  & 0.485  & 0.448  & \textbf{0.440 } & 0.703  & 0.565   \\
			& 336   & \underline{0.495 } & \underline{0.472 } & 0.517  & 0.485  & \textbf{0.452}  & \textbf{0.447 } & 0.898  & 0.641  \\
			& 720   & \textbf{0.536 } & \underline{0.517 } & 0.721  & 0.582  & \underline{0.585 } & \textbf{0.491 } & 0.948  & 0.671   \\
			& Avg   & \textbf{0.473 } & \underline{0.465 } & 0.544  & 0.496  & 0.477  & \textbf{0.451 } & 0.784  & 0.596    \\
			\midrule
			\multirow{5}[2]{*}{ETTm2} & 96    & 0.211  & 0.289  & \textbf{0.181 } & \textbf{0.264 } & \underline{0.205 } & \underline{0.277 } & 0.265  & 0.339   \\
			& 192   & \underline{0.266 } & \underline{0.324 } & \textbf{0.240 } & \textbf{0.305 } & 0.267  & 0.336  & 0.310  & 0.362    \\
			& 336   & 0.324  & 0.361  & \textbf{0.286 } & \textbf{0.336 } & \underline{0.309 } & \underline{0.347 } & 0.373  & 0.399   \\
			& 720   & \underline{0.414 } & \underline{0.414 } & \textbf{0.376 } & \textbf{0.395 } & 0.448  & 0.432  & 0.478  & 0.454  \\
			& Avg   & \underline{0.304 } & \underline{0.347 } & \textbf{0.271 } & \textbf{0.325 } & 0.307  & 0.348  & 0.356  & 0.388  \\
			\bottomrule
		\end{tabular}%
	}
	\caption{Few-shot forecasting task on 5\% training data. Results are averaged across different prediction lengths {96, 192, 336, 720}. \textbf{Bold}: best, \underline{Underline}: second-best.}
	\label{tab:few}%
\end{table}%

\noindent \textbf{Short-Term Forecasting.} This is commonly applied to tasks such as demand planning. We evaluate our model using the M4 competition dataset, which consists of 100,000 time series with 6 different frequencies, ranging from hourly to yearly. This allows for a comprehensive evaluation across varying time resolutions, as shown in Table~\ref{tab:short}. We compare our work with representative studies currently conducting short-time series forecasting on the M4 dataset. Extensive results indicate that, on average, our model achieves the lowest values for the SMAPE and OWA metrics, while the MASE metric ranks second-best.
\begin{table}
	\centering
	\scalebox{0.570}{
    \begin{tabular}{cc|c|c|c|c|c|c}
	\toprule
	\multicolumn{2}{c|}{\multirow{2}[2]{*}{Models}} & \multicolumn{1}{c|}{\multirow{2}[2]{*}{APRNet}} & \multicolumn{1}{c|}{TimeMixer++} & \multicolumn{1}{c|}{DECA} & \multicolumn{1}{c|}{Time-VLM} & \multicolumn{1}{c|}{FilterNet} & S2IP-LLM \\
	\multicolumn{2}{c|}{} & \multicolumn{1}{c|}{} & \multicolumn{1}{c|}{2025} & \multicolumn{1}{c|}{2025} & \multicolumn{1}{c|}{2025} & \multicolumn{1}{c|}{2024} & 2024 \\
	\midrule
	\multirow{3}[2]{*}{Yearly} & SMAPE & \underline{13.378 } & 13.397  & \textbf{13.288 } & 13.419  & 13.556  & 13.413  \\
	& MASE  & 3.030  & \underline{2.990 } & \textbf{2.974 } & 3.005  & 3.047  & 3.024  \\
	& OWA   & 0.791  & 0.786 & \underline{0.781 } & 0.789  & \textbf{0.770}  & 0.792  \\
	\midrule
	\multirow{3}[2]{*}{Quarterly} & SMAPE & \textbf{10.015 } & 10.206  & \multicolumn{1}{c}{\underline{10.037 }} & 10.110  & 10.348  & 10.352  \\
	& MASE  & \textbf{1.167 } & 1.201  & \underline{1.174 } & 1.178  & 1.220  & 1.228  \\
	& OWA   & \textbf{0.880 } & 0.901  & \underline{0.884 } & 0.889  & 0.915  & 0.922  \\
	\midrule
	\multirow{3}[2]{*}{Monthly} & SMAPE & \textbf{12.693 } & \underline{12.720 } & 12.762  & 12.980  & 12.905  & 12.995  \\
	& MASE  & \textbf{0.929 } & \underline{0.943 } & 0.947  & 0.963  & 0.970  & 0.970  \\
	& OWA   & \textbf{0.877 } & \underline{0.884 } & 0.897  & 0.903  & 0.905  & 0.910  \\
	\midrule
	\multirow{3}[2]{*}{Others} & SMAPE & 4.878  & \textbf{4.593 } & \underline{4.761 } & 4.795  & 5.300  & 4.805  \\
	& MASE  & \textbf{3.164 } & 3.380  & 3.207  & \underline{3.178 } & 3.799  & 3.247  \\
	& OWA   & 1.012  & 1.054  & \underline{1.007 } & \textbf{1.006 } & 1.157  & 1.017  \\
	\midrule
	\multirow{3}[2]{*}{Average} & SMAPE & \textbf{11.817 } & 11.884  & \underline{11.828 } & 11.983  & 12.082  & 12.021  \\
	& MASE  & \underline{1.581 } & 1.597  & \textbf{1.580 } & 1.595  & 1.649  & 1.612  \\
	& OWA   & \textbf{0.849 } & 0.856  & \underline{0.850 } & 0.859  & 0.877  & 0.857  \\
	\bottomrule
	\end{tabular}%
}
\caption{Full results of short-term time series forecasting on M4, with prediction horizons ranging from [6, 48]. The "Others" category includes periods with weekly, daily, and hourly. The last three rows show weighted averages across all datasets for different sampling intervals.}
	\label{tab:short}%
\end{table}%

\noindent \textbf{Zero-Shot Forecasting.} To further assess the generalization capability of APRNet, we conducted zero-shot comparative experiments with current research, as shown in Table~\ref{tab:zero}. Zero-shot forecasting requires the model to make predictions without access to historical data of the target series. When there exist discrepancies in time-series characteristics between training and test sets, it rigorously evaluates the model's capability for knowledge transfer. The experiments were primarily based on the ETT dataset. In the zero-shot tests from ETTh to ETTm and from ETTm to ETTh, the model was required to traverse time domains at the minute and hour levels, posing a significant challenge to its generalization ability. our method achieves the best overall performance. It demonstrates the second-best performance in the generalization from ETTh1 to ETTh2 and from ETTm2 to ETTm1. Notably, in terms of generalization from ETTh2 to ETTm2, the MSE metric of APRNet improved by 13.95\% compared to the second-best result.
\begin{table*}[h]
	\centering
	\scalebox{0.620}{
		\begin{tabular}{c|c|cc|cc|cc|cc|cc|cc|cc|cc|cc}
			\toprule
			& Models & \multicolumn{2}{c|}{APRNet} & \multicolumn{2}{c|}{TimeMixer++} & \multicolumn{2}{c|}{CycleNet} & \multicolumn{2}{c|}{FilterNet} & \multicolumn{2}{c|}{S2IP-LLM} & \multicolumn{2}{c|}{Time-LLM} & \multicolumn{2}{c|}{GPT4TS} & \multicolumn{2}{c|}{iTransformer} & \multicolumn{2}{c}{Dlinear} \\
			\midrule
			Datasets& Mertics & MSE   & MAE   & MSE   & MAE   & MSE   & MAE   & MSE   & MAE   & MSE   & MAE   & MSE   & MAE   & MSE   & MAE   & MSE   & MAE   & MSE   & MAE \\
			\multirow{5}[2]{*}{ETTh1-ETTh2} & 96    & \textbf{0.286 } & \textbf{0.343 } & \textbf{0.286 } & 0.345  & 0.323  & 0.368  & 0.288  & \underline{0.344 } & 0.315  & 0.377  & 0.324  & 0.368  & 0.335  & 0.374  & 0.353  & 0.394  & 0.347  & 0.400  \\
			& 192   & \textbf{0.348 } & \textbf{0.382 } & 0.376  & 0.397  & 0.376  & 0.399  & \textbf{0.348 } & \textbf{0.382 } & 0.402  & 0.407  & 0.398  & \underline{0.396 } & 0.412  & 0.417  & 0.437  & 0.445  & 0.447  & 0.460  \\
			& 336   & \underline{0.372 } & \underline{0.404 } & 0.402  & 0.421  & 0.396  & 0.421  & \textbf{0.366 } & \textbf{0.398 } & 0.453  & 0.432  & 0.410  & 0.423  & 0.441  & 0.444  & 0.482  & 0.476  & 0.515  & 0.505  \\
			& 720   & \underline{0.382 } & \underline{0.423 } & 0.439  & 0.445  & 0.402  & 0.436  & \textbf{0.372 } & \textbf{0.411 } & 0.442  & 0.451  & 0.403  & 0.449  & 0.438  & 0.452  & 0.556  & 0.506  & 0.665  & 0.589  \\
			& Avg   & \underline{0.347 } & \underline{0.388 } & 0.367  & 0.391  & 0.374  & 0.406  & \textbf{0.344 } & \textbf{0.384 } & 0.403  & 0.417  & 0.384  & 0.409  & 0.406  & 0.422  & 0.457  & 0.455  & 0.493  & 0.488  \\
			\midrule
			\multirow{5}[2]{*}{ETTh1-ETTm2} & 96    & 0.215  & 0.307  & \textbf{0.211 } & \textbf{0.303 } & 0.238  & 0.322  & \underline{0.213 } & \underline{0.305 } & 0.242  & 0.319  & 0.236  & 0.320  & 0.236  & 0.315  & 0.247  & 0.319  & 0.255  & 0.357  \\
			& 192   & \underline{0.267 } & \underline{0.337 } & 0.286  & 0.348  & 0.287  & 0.354  & \underline{0.267 } & \textbf{0.336 } & 0.286  & \underline{0.337 } & \textbf{0.265 } & 0.353  & 0.287  & 0.342  & 0.293  & 0.350  & 0.338  & 0.413  \\
			& 336   & \textbf{0.317 } & \textbf{0.366 } & 0.351  & 0.386  & 0.335  & 0.372  & \textbf{0.317 } & \underline{0.367 } & 0.351  & \underline{0.367 } & 0.337  & 0.376  & 0.341  & 0.374  & 0.364  & 0.419  & 0.425  & 0.465  \\
			& 720   & \textbf{0.403 } & \textbf{0.413 } & 0.448  & 0.439  & 0.415  & 0.418  & \underline{0.405 } & 0.417  & 0.422  & \underline{0.416 } & 0.429  & 0.430  & 0.435  & 0.422  & 0.534  & 0.470  & 0.640  & 0.573  \\
			& Avg   & \textbf{0.301 } & \textbf{0.356 } & \textbf{0.301 } & \underline{0.357 } & 0.319  & 0.367  & \textbf{0.301 } & \textbf{0.356 } & 0.325  & 0.360  & 0.317  & 0.370  & 0.325  & 0.363  & 0.360  & 0.390  & 0.415  & 0.452  \\
			\midrule
			\multirow{5}[2]{*}{ETTh2-ETTh1} & 96    & \textbf{0.437 } & \textbf{0.447 } & \underline{0.446 } & \underline{0.450}  & 0.504  & 0.487 & 0.640  & 0.554  & 0.668  & 0.567  & 0.618  & 0.515  & 0.732  & 0.577  & 0.854  & 0.606  & 0.689  & 0.555  \\
			& 192   & \textbf{0.465 } & \textbf{0.466 } & 0.572  & 0.529  & \underline{0.539 } & \underline{0.509 } & 0.714  & 0.592  & 0.575  & 0.526  & 0.715  & 0.570  & 0.758  & 0.559  & 0.863  & 0.615  & 0.707  & 0.568  \\
			& 336   & \underline{0.525 } & \textbf{0.504 } & \textbf{0.502 } & \textbf{0.504 } & 0.569  & 0.530  & 0.631  & 0.561  & 0.655  & 0.577  & 0.636  & \underline{0.523 } & 0.759  & 0.578  & 0.867  & 0.626  & 0.710  & 0.577  \\
			& 720   & \underline{0.626 } & 0.566  & \textbf{0.565 } & \textbf{0.530 } & 0.634  & 0.574  & 0.768  & 0.623  & 0.778  & 0.568  & 0.683  & \underline{0.553 } & 0.781  & 0.597  & 0.887  & 0.654  & 0.704  & 0.596  \\
			& Avg   & \textbf{0.513 } & \textbf{0.496 } & \underline{0.521 } & \underline{0.503 } & 0.562  & 0.525  & 0.688  & 0.583  & 0.669  & 0.560  & 0.663  & 0.540  & 0.757  & 0.578  & 0.868  & 0.625  & 0.703  & 0.574  \\
			\midrule
			\multirow{5}[2]{*}{ETTh2-ETTm2} & 96    & \underline{0.220 } & 0.311  & 0.254  & 0.328  & \textbf{0.217 } & \underline{0.308 } & 0.236  & 0.323  & 0.221  & \textbf{0.303 } & 0.258  & 0.326  & 0.253  & 0.329  & 0.244  & 0.330  & 0.240  & 0.336  \\
			& 192   & \textbf{0.262 } & \textbf{0.333 } & 0.291  & 0.352  & \underline{0.268 } & \underline{0.338 } & 0.275  & 0.343  & 0.295  & 0.344  & 0.303  & 0.342  & 0.293  & 0.346  & 0.291  & 0.356  & 0.295  & 0.369  \\
			& 336   & \textbf{0.310 } & \textbf{0.361 } & 0.387  & 0.403  & \underline{0.314 } & \underline{0.365 } & 0.322  & 0.372  & 0.340  & 0.376  & 0.356  & 0.383  & 0.347  & 0.376  & 0.351  & 0.391  & 0.345  & 0.397  \\
			& 720   & \textbf{0.393 } & \textbf{0.409 } & 0.435  & 0.436  & \underline{0.399 } & \underline{0.413 } & 0.419  & 0.431  & 0.453  & 0.428  & 0.440  & 0.434  & 0.446  & 0.429  & 0.452  & 0.451  & 0.432  & 0.442  \\
			& Avg   & \textbf{0.296 } & \textbf{0.354 } & 0.342  & 0.380  & \underline{0.300 } & \underline{0.356 } & 0.313  & 0.367  & 0.327  & 0.363  & 0.339  & 0.371  & 0.335  & 0.370  & 0.335  & 0.382  & 0.328  & 0.386  \\
			\midrule
			\multirow{5}[2]{*}{ETTm1-ETTh2} & 96    & \textbf{0.334 } & \textbf{0.379 } & 0.363  & 0.400  & 0.374  & 0.401  & \underline{0.342 } & 0.389  & 0.358  & \underline{0.382 } & 0.355  & 0.403  & 0.353  & 0.392  & 0.371  & 0.407  & 0.365  & 0.415  \\
			& 192   & \textbf{0.399 } & \textbf{0.415 } & 0.415  & 0.424  & 0.435  & 0.438  & \underline{0.405 } & \underline{0.420 } & 0.454  & 0.444  & 0.449  & 0.450  & 0.443  & 0.437  & 0.463  & 0.458  & 0.454  & 0.462  \\
			& 336   & 0.419  & \underline{0.432 } & 0.438  & 0.451  & \textbf{0.394 } & \textbf{0.423 } & \underline{0.413 } & 0.433  & 0.488  & 0.452  & 0.479  & 0.467  & 0.469  & 0.461  & 0.481  & 0.485  & 0.496  & 0.494  \\
			& 720   & 0.436  & 0.456  & 0.437  & 0.454  & \textbf{0.417 } & \textbf{0.445 } & \underline{0.429 } & \underline{0.453 } & 0.469  & 0.478  & 0.477  & 0.476  & 0.466  & 0.468  & 0.503  & 0.482  & 0.541  & 0.529  \\
			& Avg   & \textbf{0.397 } & \textbf{0.421 } & 0.413  & 0.432  & \underline{0.405 } & 0.427  & \textbf{0.397 } & \underline{0.424 } & 0.442  & 0.439  & 0.440  & 0.449  & 0.433  & 0.439  & 0.455  & 0.458  & 0.464  & 0.475  \\
			\midrule
			\multirow{5}[2]{*}{ETTm1-ETTm2} & 96    & \textbf{0.179 } & \textbf{0.263 } & 0.189  & 0.274  & 0.197  & 0.282  & \underline{0.182 } & \underline{0.267 } & 0.203  & 0.299  & 0.218  & 0.271  & 0.217  & 0.294  & 0.219  & 0.305  & 0.221  & 0.314  \\
			& 192   & \textbf{0.234 } & \textbf{0.300 } & 0.240  & \underline{0.304 } & 0.266  & 0.324  & \underline{0.238 } & \underline{0.304 } & 0.272  & 0.325  & 0.288  & 0.335  & 0.277  & 0.327  & 0.277  & 0.347  & 0.286  & 0.359  \\
			& 336   & \underline{0.289 } & \underline{0.336 } & 0.298  & 0.344  & \textbf{0.281 } & \textbf{0.332 } & \underline{0.289 } & 0.337  & 0.303  & 0.347  & 0.322  & 0.355  & 0.331  & 0.360  & 0.354  & 0.378  & 0.357  & 0.406  \\
			& 720   & \textbf{0.365 } & \textbf{0.383 } & 0.372  & \underline{0.386 } & \textbf{0.365 } & 0.384  & \underline{0.368 } & \underline{0.386 } & 0.436  & 0.418  & 0.414  & 0.409  & 0.429  & 0.413  & 0.426  & 0.420  & 0.476  & 0.476  \\
			& Avg   & \textbf{0.267 } & \textbf{0.321 } & 0.275  & 0.327  & 0.277  & 0.331  & \underline{0.269 } & \underline{0.324 } & 0.304  & 0.347  & 0.311  & 0.343  & 0.313  & 0.348  & 0.319  & 0.363  & 0.335  & 0.389  \\
			\midrule
			\multirow{5}[2]{*}{ETTm2-ETTh2} & 96    & \textbf{0.303 } & \textbf{0.362 } & 0.381 & 0.416 & \underline{0.318 } & \underline{0.376 } & 0.347  & 0.392  & 0.324  & 0.383  & 0.334  & 0.416  & 0.360  & 0.401  & 0.347  & 0.401  & 0.333  & 0.391  \\
			& 192   & \underline{0.384 } & \textbf{0.411 } & 0.412  & \underline{0.422 } & \textbf{0.382 } & \textbf{0.411 } & 0.406  & 0.425  & 0.403  & \underline{0.422 } & 0.439  & 0.441  & 0.434  & 0.437  & 0.438  & 0.444  & 0.441  & 0.456  \\
			& 336   & \textbf{0.413 } & \textbf{0.432 } & 0.463  & 0.460  & \textbf{0.413 } & \underline{0.434 } & 0.445  & 0.454  & \underline{0.434 } & 0.442  & 0.455  & 0.457  & 0.460  & 0.459  & 0.459  & 0.464  & 0.505  & 0.503  \\
			& 720   & 0.436  & 0.456  & \underline{0.434 } & \underline{0.454 } & \textbf{0.426 } & \textbf{0.452 } & 0.479  & 0.482  & 0.462  & 0.467  & 0.488  & 0.479  & 0.485  & 0.477  & 0.485  & 0.477  & 0.543  & 0.534  \\
			& Avg   & \textbf{0.384 } & \textbf{0.415 } & 0.423  & 0.438  & \underline{0.385 } & \underline{0.418 } & 0.419  & 0.438  & 0.406  & 0.429  & 0.429  & 0.448  & 0.435  & 0.443  & 0.432  & 0.447  & 0.455  & 0.471  \\
			\midrule
			\multirow{5}[2]{*}{ETTm2-ETTm1} & 96    & \textbf{0.436 } & \textbf{0.431 } & \underline{0.444 } & \underline{0.433 } & 0.472  & 0.452  & 0.496  & 0.462  & 0.583  & 0.524  & 0.488  & 0.445  & 0.747  & 0.558  & 0.619  & 0.564  & 0.570  & 0.490  \\
			& 192   & \textbf{0.503 } & \underline{0.472 } & \underline{0.506 } & \textbf{0.469 } & 0.522  & 0.480  & 0.613  & 0.523  & 0.609  & 0.501  & 0.555  & 0.464  & 0.781  & 0.560  & 0.685  & 0.565  & 0.590  & 0.506  \\
			& 336   & \textbf{0.512 } & \textbf{0.480 } & 0.585  & 0.513  & \underline{0.547 } & \underline{0.497 } & 0.621  & 0.530  & 0.585  & 0.522  & 0.608  & 0.538  & 0.778  & 0.578  & 0.792  & 0.578  & 0.706  & 0.567  \\
			& 720   & 0.613  & 0.534  & \underline{0.605 } & \underline{0.525 } & \textbf{0.597 } & \textbf{0.524 } & 0.707  & 0.571  & 0.712  & 0.579  & 0.699  & 0.566  & 0.769  & 0.573  & 0.727  & 0.579  & 0.731  & 0.584  \\
			& Avg   & \textbf{0.516 } & \textbf{0.479 } & \underline{0.535 } & \underline{0.485 } & \underline{0.535 } & 0.488  & 0.609  & 0.522  & 0.622  & 0.532  & 0.588  & 0.503  & 0.769  & 0.567  & 0.706  & 0.572  & 0.649  & 0.537  \\
			\bottomrule
		\end{tabular}%
	
	}
		\caption{Full results of Zero-shot learning: the first column X → Y indicates training on dataset X and testing on dataset Y.}	\label{tab:zero}%
\end{table*}%

\subsection{Ablation Study}
We explore the two key components of our APRNet, including the joint modeling of channels and sequences, as well as the learning of local stationary information regarding amplitude and phase. We selected two representative datasets, weather and electricity. The weather dataset has a feature dimension of 21, while the electricity dataset has a feature dimension of 321, with similar sequence lengths, making them suitable for observing how changes in feature dimensions affect the key components' relationships. 

\noindent \textbf{Sequence and Channel Modeling.}
Table~\ref{tab:ablation1} lists the ablation experiments on sequence and channel modeling. Modeling in the sequence dimension captures temporal dependencies and dynamic patterns within the data, demonstrating superior performance over channel-dimension modeling in the weather dataset. However, due to the rich channel information in the electricity dataset, modeling solely in the sequence dimension can lead to the loss of feature information, resulting in poorer detection performance. By decoupling channels and sequences, our method achieves improved performance on both the weather and electricity datasets.
\begin{table}[H]
	\centering
	\scalebox{0.650}{
		\begin{tabular}{c|c|cc|cc}
			\toprule
			\multicolumn{2}{c|}{Datasets} & \multicolumn{2}{c|}{Weather} & \multicolumn{2}{c}{Electricity} \\
			\midrule
			Channel & Sequence & MSE   & MAE   & MSE   & MAE \\
			\midrule
			\checkmark&       & 0.231 & 0.267 & \underline{0.158} & \underline{0.252} \\
			&     \checkmark  & \underline{0.230}  & \textbf{0.244} & 0.164 & 0.263 \\
			\checkmark    &  \checkmark     & \textbf{0.222} & \underline{0.263} & \textbf{0.156} & \textbf{0.251} \\
			\bottomrule
		\end{tabular}%
		
	}
	\caption{The ablation experiments on sequences and channels present the average results across step lengths of $\{96, 192, 336, 720\}$.}\label{tab:ablation1}%
\end{table}%

\noindent \textbf{Amplitude and Phase Modeling.}
Table~\ref{tab:ablation2} lists the ablation implementation of our approach to learning amplitude and phase across multiple frequency groups to capture local relationships. Amplitude focuses more on periodic, trend, and abrupt information, demonstrating superior forecasting performance compared to learning within the phase domain. Phase encompasses the temporal offset of signals in time series. By integrating these two critical factors, we mine local stationary information across different frequencies, thereby enhancing the model's predictive performance.
\begin{table}
	\centering
	\scalebox{0.650}{
		\begin{tabular}{c|c|cc|cc}
			\toprule
			\multicolumn{2}{c|}{Datasets} & \multicolumn{2}{c|}{Weather} & \multicolumn{2}{c}{Electricity} \\
			\midrule
			Amplitude & Phase &    MSE   &   MAE    &   MSE    & MAE \\
			\midrule
			\checkmark   &       &\underline{0.223} & \textbf{0.263} & \underline{0.160}  & 0.285 \\
			&    \checkmark   & 0.228 & \underline{0.267} & 0.165 & \underline{0.261} \\
			\checkmark   &  \checkmark     & \textbf{0.222} & \textbf{0.263} & \textbf{0.156} & \textbf{0.251} \\
			\bottomrule
		\end{tabular}%
	
	}
		\caption{The ablation experiments on amplitude and phase present the average results across step lengths of $\{96, 192, 336, 720\}$}	\label{tab:ablation2}%
\end{table}%

\begin{figure}
	\centering
	\includegraphics[width=\linewidth]{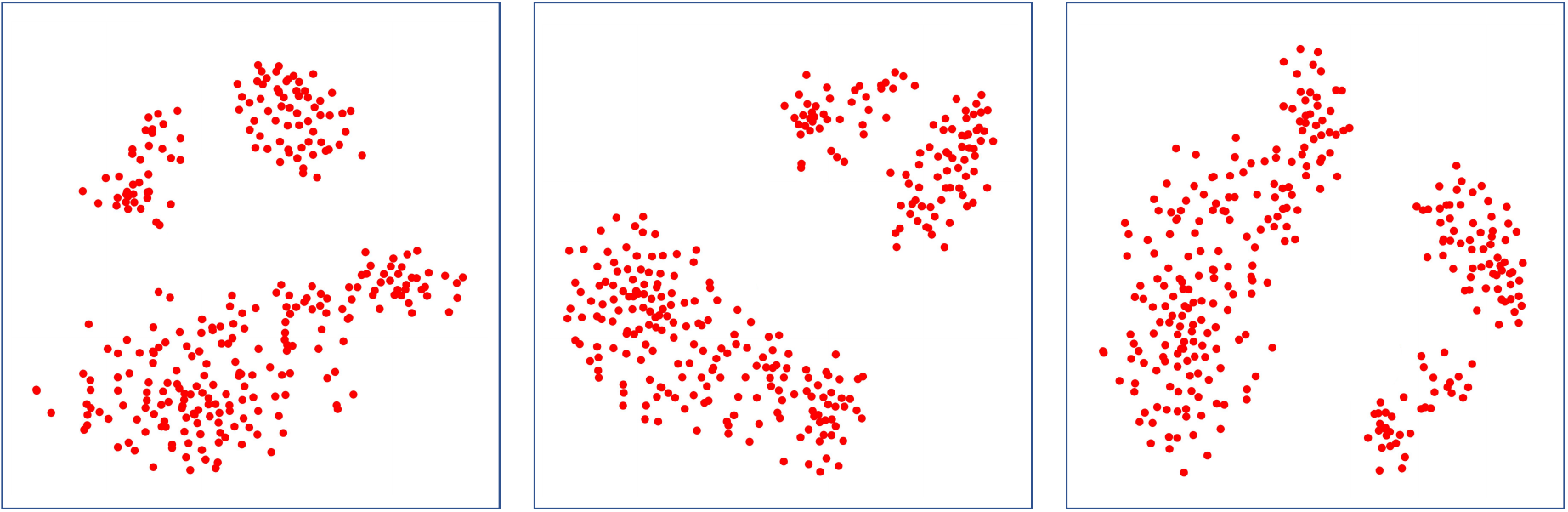}
	\caption{t-SNE visualization through APLC analysis.}
	\label{t-SNE}
\end{figure}

\noindent \textbf{Parameter Sensitivity.}
We illustrate the capture of stationary information following analysis by the APLC module in Fig~\ref{t-SNE}. The three t-SNE results depicted in the figure demonstrate the clustering phenomenon of stationary segments we observed in the feature space. The model can effectively capture stationary information, and then the data points corresponding to the stationary information will form tightly clustered shapes in low-dimensional space. By carefully comparing the visualized results output by different models, we can intuitively evaluate each model's ability to extract stationary features. Specifically, the higher the degree of clustering among the data points and the clearer the boundaries between clusters, the more thoroughly the model has explored the stationary patterns. 

\begin{figure}[H]
	\centering
	\includegraphics[width=0.90\linewidth]{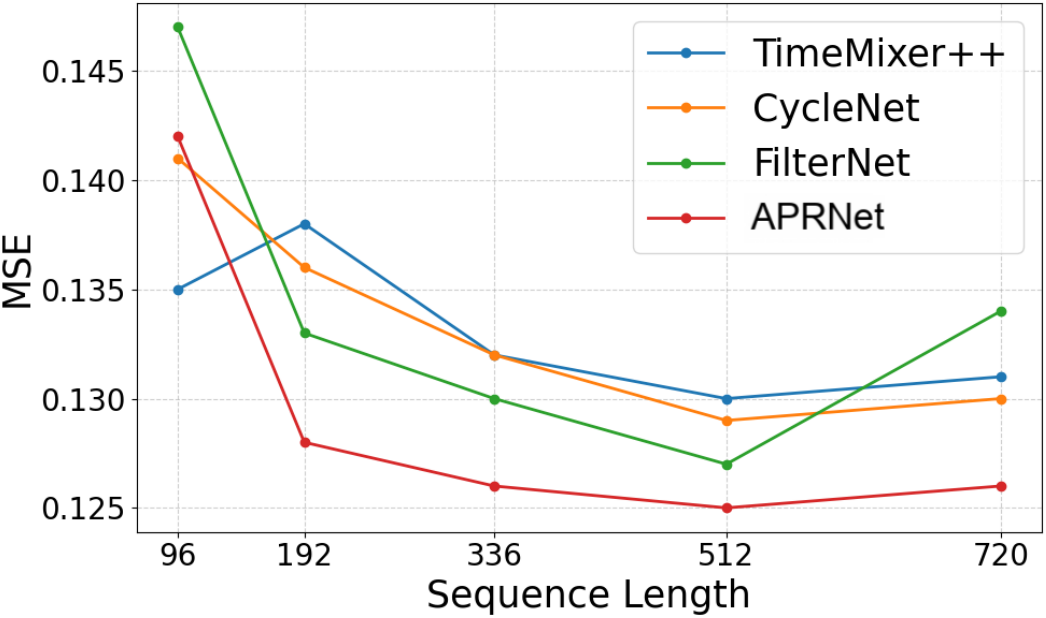}
	\caption{The experiments were conducted on the electricity dataset, with input sequence lengths set to $\{96, 192, 336, 512, 720\}$. The forecasting horizon was 96.}
	\label{Diff}
\end{figure}

\noindent \textbf{Input Sequence.}
Expanding the look-back window can provide more information for forecasting the future time steps, potentially leading to improvements in prediction performance. However, different algorithms have varying capacities to comprehend input sequence lengths. This paper conducts comparisons under the condition of an input length of 512. To further verify the impact of different sequence lengths on model performance, we carried out comparative experiments, the results of which are shown in Fig.~\ref{Diff}. As the look-back window increases, the error of APRNet continuously decreases, indicating that our method can effectively capture and summarize temporal information.

\noindent \textbf{KLC Module.}
This module employs the KAN linear model to learn the dependency relationships between amplitude and phase. To verify the effectiveness of KAN, we conduct ablation experiments by replacing KAN with Linear layers and convolutional layers, as shown in Table~\ref{tab:kan}. The experimental results demonstrate that KAN approximates complex functions by combining nonlinear functions, making it more suitable for capturing relationships among frequencies.
\begin{table}[htbp]
	\centering
	\scalebox{0.680}{
	\begin{tabular}{cc|cc|cc}
		\toprule
		\multicolumn{2}{c|}{Datasets} & \multicolumn{2}{c|}{Electricity} & \multicolumn{2}{c}{Traffic} \\
		\midrule
		\multicolumn{2}{c|}{Metrics} & MSE   & MAE   & MSE   & MAE \\
		\midrule
		\multicolumn{2}{c|}{convolution}  & 0.165 & 0.264 & 0.408 & 0.282 \\
		\multicolumn{2}{c|}{Linear}  & \underline{0.160}  & \underline{0.259} & \underline{0.390} & \underline{0.273} \\
		\multicolumn{2}{c|}{KAN} & \textbf{0.156} & \textbf{0.251} & \textbf{0.389} & \textbf{0.271} \\
		\bottomrule
	\end{tabular}%
}
	
	\caption{KAN-based ablation results, the average results across step lengths of $\{96, 192, 336, 720\}$}
	\label{tab:kan}%
\end{table}%
\subsection{Efficiency Analysis}
Here, we compared APRNet with TimeMixer++~\cite{TimeXier} and representative research works in terms of Multiply-Accumulate Operations (MACs) to verify that APRNet is a lightweight and efficient architecture. In the same input sequence length and experimental parameters, it is evident that APRNet demonstrates significant advantages in terms of MACs.
\begin{table}[H]
	\centering

	\scalebox{0.600}{
		\begin{tabular}{c|c|c|c|c|c|c|c}
			\toprule
			Datasets & ETTh1 & ETTh2 & ETTm1 & ETTm2 & Weather & Electricity & Traffic \\
			\midrule
			iTransformer & 77.46M & 19.86M & 19.86M & 19.86M & 1.16G & 16.29G & 43.8G \\
			TimeMixer++ & 98.92M & 98.92M & 98.92M & 197.86M & 296.78M & 4.53G & 130.29G \\
			APRNet & 133.32M & 33.47M & 45.34M & 14.69M & 135.11M & 2.09G & 17.78G \\
			\bottomrule
		\end{tabular}%
	}

	\caption{The comparison results with MACs}	\label{tab:efficiency}%
\end{table}%
\section{Conclusion}
In this paper, we introduced the Amplitude-Phase Reconstruct Network (APRNet) for time series forecasting. This method explores stationarity from the perspectives of amplitude and phase in the frequency domain for time series forecasting. The Amplitude-Phase Local Correlation (APLC) module models the local relationships between amplitude and phase. It captures dynamic characteristics in the frequency domain by restructuring frequency signals, enhancing its ability to forecast future state performance. Experimental results reveal that APRNet not only surpasses existing state-of-the-art (SOTA) methods but also demonstrates remarkable adaptability to multi-task scenarios, highlighting its capability to effectively process non-stationary information and capture fine-grained details while preserving operational efficiency.
\newpage
\bibliography{aaai2026}

\end{document}